\documentclass[conference]{IEEEtran}
\usepackage[left=0.75in,right=0.75in,top=1.0in,bottom=0.75in]{geometry}
\IEEEoverridecommandlockouts

\usepackage{cite}
\usepackage{amsmath,amssymb,amsfonts}
\usepackage{algorithmic}
\usepackage{graphicx}
\usepackage{textcomp}
\usepackage{xcolor}
\def\BibTeX{{\rm B\kern-.05em{\sc i\kern-.025em b}\kern-.08em
    T\kern-.1667em\lower.7ex\hbox{E}\kern-.125emX}}

\usepackage[utf8]{inputenc} %
\usepackage[T1]{fontenc}    %
\usepackage{hyperref}       %
\usepackage{url}            %
\usepackage{booktabs}       %
\usepackage{amsfonts}       %
\usepackage{nicefrac}       %
\usepackage{microtype}      %
\usepackage{xcolor}         %
\usepackage{tabularx, makecell, multirow} 
\usepackage{amsmath,amssymb,multicol,latexsym}
\usepackage{pgfplots,pgfplotstable}
\pgfplotsset{compat=1.16}
\usepackage{tikz}
\usepackage{hyperref}
\usepackage{cleveref}
\usepackage[font=footnotesize]{caption}
\usepackage[font=footnotesize]{subcaption}

\DeclareMathAlphabet{\mathcal}{OMS}{cmsy}{m}{n}

\usepackage{url}
\usepackage{graphicx}
\graphicspath{{figures/}}
\usepackage{cite}
\usepackage{flushend} %

\usepackage[colorinlistoftodos,prependcaption,textsize=small]{todonotes}
\usepackage{regexpatch}
\makeatletter
\xpatchcmd{\@todo}{\setkeys{todonotes}{#1}}{\setkeys{todonotes}{inline,#1}}{}{}

\graphicspath{{figures/}{figures/ssg/}{figures/titleimg/}}

\begin{document}

\title{Heterogeneous Graph-based Trajectory Prediction using Local Map Context and Social Interactions}

\author{
Daniel~Grimm$^{1,2}$,
Maximilian~Zipfl$^{1,2}$,
Felix Hertlein$^{1,2}$,
Alexander Naumann$^{1,2}$,
Juergen Luettin$^{3}$,\\
Steffen Thoma$^{1}$,
Stefan Schmid$^{3}$,
Lavdim Halilaj$^{3}$,
Achim Rettinger$^{1}$,
and J.~Marius~Zöllner$^{1,2}$
\thanks{$^{1}$FZI Research Center for Information Technology, Karlsruhe, Germany
{\tt\small \{grimm, zipfl, zoellner\}@fzi.de}}%
\thanks{$^{2}$Karlsruhe Institute of Technology, Karlsruhe, Germany}%
\thanks{$^{3}$Bosch Center for Artificial Intelligence, Renningen, Germany}%

}%

\maketitle

\begin{abstract}

Precisely predicting the future trajectories of surrounding traffic participants is a crucial but challenging problem in autonomous driving, due to complex interactions between traffic agents, map context and traffic rules.
Vector-based approaches have recently shown to achieve among the best performances on trajectory prediction benchmarks.
These methods model simple interactions between traffic agents but don't distinguish between relation-type and attributes like their distance along the road.
Furthermore, they represent lanes only by sequences of vectors representing center lines and ignore context information like lane dividers and other road elements.
We present a novel approach for vector-based trajectory prediction that addresses these shortcomings by leveraging three crucial sources of information:
First, we model interactions between traffic agents by a semantic scene graph, that accounts for the nature and important features of their relation.
Second, we extract agent-centric image-based map features to model the local map context.
Finally, we generate anchor paths to enforce the policy in multi-modal prediction to permitted trajectories only.
Each of these three enhancements shows advantages over the baseline model HoliGraph \cite{holigraph}.
\end{abstract}


\date{May 2023}

\maketitle

\section{Introduction}
\label{sec:introduction}
Autonomous vehicles (AVs) have a great potential to change the future of public transportation to the benefit of our safety and the environment.
To realize this potential, AVs must be able to handle all potential traffic scenarios, which requires them to detect all surrounding traffic participants and predict their movements accurately.
However, motion prediction in autonomous driving is a hard task, mainly because of the unknown intention of other traffic participants, their complex interactions, the complex nature of driving scene context like lanes, lane dividers, pedestrian crossings, as well as the governing traffic rules.
Therefore, state-of-the-art models often predict multiple trajectories per traffic participant to capture different possible driving behaviors \cite{Zhou2022HiVTHV}.
However one cannot explicitly define what those behaviors are and therefore it is hard to determine, if a prediction matches the learned driving behavior or if it is out of distribution.
\begin{figure}[t]
  \centering
  \def\svgwidth{0.88\columnwidth}
  \scriptsize{
\begingroup%
  \makeatletter%
  \providecommand\color[2][]{%
    \errmessage{(Inkscape) Color is used for the text in Inkscape, but the package 'color.sty' is not loaded}%
    \renewcommand\color[2][]{}%
  }%
  \providecommand\transparent[1]{%
    \errmessage{(Inkscape) Transparency is used (non-zero) for the text in Inkscape, but the package 'transparent.sty' is not loaded}%
    \renewcommand\transparent[1]{}%
  }%
  \providecommand\rotatebox[2]{#2}%
  \newcommand*\fsize{\dimexpr\f@size pt\relax}%
  \newcommand*\lineheight[1]{\fontsize{\fsize}{#1\fsize}\selectfont}%
  \ifx\svgwidth\undefined%
    \setlength{\unitlength}{371.22936819bp}%
    \ifx\svgscale\undefined%
      \relax%
    \else%
      \setlength{\unitlength}{\unitlength * \real{\svgscale}}%
    \fi%
  \else%
    \setlength{\unitlength}{\svgwidth}%
  \fi%
  \global\let\svgwidth\undefined%
  \global\let\svgscale\undefined%
  \makeatother%
  \begin{picture}(1,0.79962869)%
    \lineheight{1}%
    \setlength\tabcolsep{0pt}%
    \put(0,0){\includegraphics[width=\unitlength,page=1]{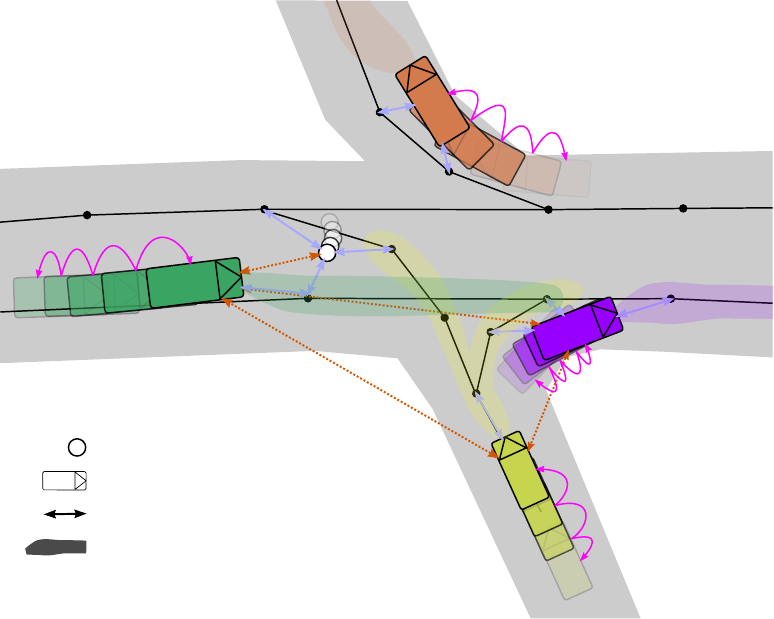}}%
    \put(0.14251365,0.21461379){\color[rgb]{0,0,0}\makebox(0,0)[lt]{\lineheight{1.25}\smash{\begin{tabular}[t]{l}Non-road-bound agent node\end{tabular}}}}%
    \put(0.14218482,0.17099533){\color[rgb]{0,0,0}\makebox(0,0)[lt]{\lineheight{1.25}\smash{\begin{tabular}[t]{l}road-bound-agent node\end{tabular}}}}%
    \put(0.14166692,0.12737694){\color[rgb]{0,0,0}\makebox(0,0)[lt]{\lineheight{1.25}\smash{\begin{tabular}[t]{l}edge\end{tabular}}}}%
    \put(0.14169157,0.08374246){\color[rgb]{0,0,0}\makebox(0,0)[lt]{\lineheight{1.25}\smash{\begin{tabular}[t]{l}anchor path\end{tabular}}}}%
    \put(0.14215194,0.25821535){\color[rgb]{0,0,0}\makebox(0,0)[lt]{\lineheight{1.25}\smash{\begin{tabular}[t]{l}map node\end{tabular}}}}%
    \put(0,0){\includegraphics[width=\unitlength,page=2]{obenrechts2_leg.pdf}}%
  \end{picture}%
\endgroup%
}
  \caption{Schematic illustration of the heterogeneous graph. It shows different node and edge types as well as anchor paths.}
  \label{fig:overview}
\end{figure}

In this work, we propose an approach that leverages so-called anchor paths. Anchor paths define road segments/lanes a traffic participant is permitted to travel on.
Therefore, the predicted trajectory with the assigned anchor path allows for the detection of potential invalid predictions, i.e. predictions that do not follow the anchor path.
Anchor paths are only reasonable for traffic participants that travel on the road, not pedestrians.
In the following we refer to traffic participants as agents.
We distinguish between road-bound and non-road-bound agents where anchor paths are only used by the former.
There are two common approaches to represent a traffic scene, a grid-based approach and a vector-based approach.
Grid-based approaches represent the traffic scene in one layered image with the agent to be predicted in the center of the image.
They usually excel in the representation of the surrounding environment and are typically used by Convolutional Neural Networks (CNNs), like \cite{djuric_uncertainty-aware_2020}.
Vector-based approaches present agents and road topology as sets of vectors.
This representation is then used in transformer-like architectures \cite{ngiam_scene_2021,Zhou2022HiVTHV} or Graph Neural Networks (GNNs) \cite{jia_hdgt_2022}.
These models naturally distinguish between different traffic participants, making it simple to predict multiple traffic participants simultaneously.
Here, we combine the benefits of the two representations into a single heterogeneous spatio-temporal graph that is schematically depicted in \Cref{fig:overview}.
Social interaction between agents is often modeled in a fully-connected approach \cite{Jia2022HDGTHD}.
In this work we use the concept of a semantic scene graph to determine potential interactions beforehand and thus reduce the number of edges in the graph without losing performance.

\section{Related Work}
\label{sec:related_work}
The first published ml-based motion prediction models used a raster-based approach that represented the scene as a spatial grid including semantic scene information, e.g., road topology, walkways etc. \cite{djuric_uncertainty-aware_2020}.
These have been extended to generate multiple possible trajectories while also estimating their probabilities \cite{cui_multimodal_2019,bansal_chauffeurnet_2018,hong_rules_2019,PhanMinh2019CoverNetMB,gilles_gohome_2021}.
Some works \cite{casas_intentnet_2021,Rhinehart2019PRECOGPC} divided motion prediction in a goal- or intention estimation followed by a trajectory completion.
An approach that uses a fixed set of state-sequence anchors learned from the training set is described in \cite{chai_multipath_2019}.
A way to learn latent representations of anchors trajectories can be found in \cite{Varadarajan2021MultiPathEI}.
Transformer-based models \cite{Tang2019MultipleFP,messaoud_attention_2021,Yuan2021AgentFormerAT,khandelwal_what-if_2020,liu_multimodal_2021,girgis_latent_2022} and 
graph-based approaches \cite{Gao2020VectorNetEH,Liang2020Learning,Zeng2021LaneRCNNDR,Deo2021MultimodalTP,Huang2021MultimodalMP} represent traffic participants and map features as sets of vectors.
In \cite{Zhao2020TNTTT, gu_densetnt_2021}, the VectorNet model \cite{Gao2020VectorNetEH} is extended to target-based prediction and anchor-free dense goal sets.
A hierarchical vector transformer based approach is proposed in \cite{Zhou2022HiVTHV} to obtain a translation-invariant scene representation and rotation-invariant spatial learning.
This is done by local context feature encoding followed by global message passing among agent-centric local regions. 
Heterogeneous graphs, i.e., graphs with different entity types, such as vehicles, bicycles or pedestrians as well as relation types, like agent-to-lane or agent-to-agent, are first introduced in  \cite{hu_heterogeneous_2020}.
Motion prediction methods for heterogeneous graphs are described in \cite{Mo2022MultiAgentTP,Jia2022HDGTHD,Monninger2023SCENE,Wonsak2022Multi}.
This work builds upon the work of HoliGraph  (HG)\cite{holigraph}, a holistic heterogeneous graph-based approach for motion prediction using an encoder-decoder architecture.
The encoder consists of a GNN while the decoder is modeled as a Multilayer Perceptron (MLP) with multiple heads.
\section{Concept}
\label{sec:concept}
\begin{figure*}[t]
    \vspace{4pt}
    \centering
    \includegraphics[width=\textwidth]{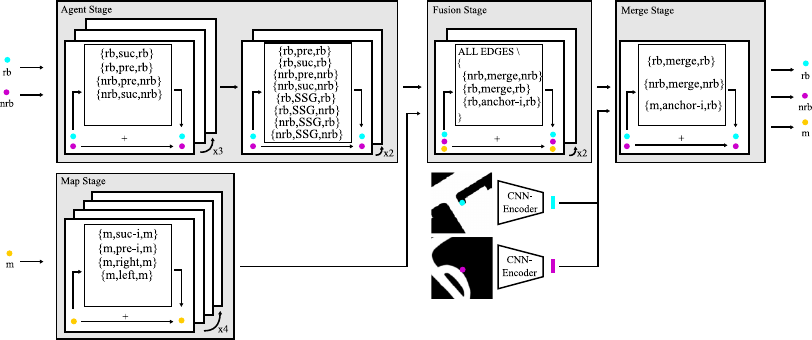}
    \caption{GNN architecture: It consists of four different stages. Each using different type of nodes and edges. \emph{Agent Stage} and \emph{Map Stage} update the respective nodes separately. In the \emph{Fusion Stage} the complete GNN is used. The \emph{Merge Stage} fuses the temporal context of each agent and outputs a feature vector for every agent in the scene.}
    \label{fig:arcitecture}
\end{figure*}
In this work, we present beneficial additions to HoliGraph \cite{holigraph}.%
Firstly, we present the overall architecture including input data and the heterogeneous graph in \Cref{sec:overview}.
Secondly, we describe the generation of a semantic scene graph that tracks potential interactions between agents and therefore replaces the fully-connected approach of the baseline in \Cref{sec:ssg}.
Thirdly, we combine our polyline-based map representation in \Cref{sec:map_encoding} with a grid-based approach to capture richer details of the HD-Map, like lane borders, lane dividers, walkways and pedestrian crossings.
Fourthly, we condition the decoder of the model to attend to anchor paths (AP), i.e., possible paths an agent can travel on, see \Cref{sec:ap}.
Lastly, we present our Loss, introducing an orientation loss $\mathcal{L}_{\text{yaw}}$ in \Cref{seq:loss}
\subsection{Overall architecture}
\label{sec:overview}
In general, our heterogeneous graph $\mathcal{G}$ consists of a set of nodes $\mathcal{N}$ and a set of edges $\mathcal{E}$.
We distinguish three different types of nodes:
\begin{itemize}
    \item rb-agent-nodes ${}^{rb}n_{i}^{t}$. Each node refers to a state at time-step $t$ of agent $i$ of type \emph{road-bound}. A state consists of the 2d-position and 2d-velocity.
    \item nrb-agent-nodes ${}^{nrb}n_{i}^{t}$. They consist of the same features as rb-agent-nodes but belong to agents of type \emph{non-road-bound} and thus are treated differently by the model.
    \item map-nodes ${}^{m}n_{i}$. A map-node describes a segment of a centerline of a lane in the HD-Map. It consists of 2d-position and 2d-direction.
\end{itemize}
Agents of type \emph{road-bound} typically travel along the centerline of roads while \emph{non-road-bound} usually don't.
The set of edges $\mathcal{E}$ contains a variety of different edge-types $e_{j,r,i}$ where $j$ and $i$ refer to the source and destination node type, respectively and $r$ specifies the relation.
Each edge except those belonging to the SSG, see \cref{sec:ssg},  has the euclidean distance and the relative cartesian coordinates between the connection nodes as features, execpt edges .
A total list of all edges is given in \Cref{tab:edges}.

In an initial \emph{Embedding Stage} all nodes and edges are embedded to a hidden dimension of $128$ with a MLP.
Similar to \cite{hu_heterogeneous_2020} a time-encoding is added to nodes that hold temporal information.
Next, in the GNN, the features of the nodes are updated according to the architecture in \Cref{fig:arcitecture}.
In the \emph{Agent Stage} and \emph{Map Stage}, respectively, nodes belonging to traffic participants and the HD-Map are updated separately.
Afterwards, information is shared across node types in the \emph{Fusion Stage}.
Finally, in the \emph{Merge Stage} a single vector with the latent features of an agent $z_{\text{agent},i}$ is generated for each agent in the traffic scene.

The trajectory decoder has multiple heads - one head for every mode - where every head consists of two small MLPs that output a trajectory and a score.
The weights are shared across agents of the same type and same mode number $k$.
The decoder is trained in a winner-takes-all mentality meaning only the weights of the head with the best prediction are updated.
\begin{table}[h]
    \centering
    \hspace{0.1pt}
    \caption{List of all edges in our model and the corresponding operation on them. eGCN refers to the edge-enhanced graph convolution defined in \cite{holigraph} and GatV2 refers to the graph attention from \cite{brody_how_2022}.}
    \begin{tabular}{@{}l c c@{}}
        \toprule
        edge name & domain & graph operation \\
        \midrule
        \{rb, suc, rb\} & temporal & eGCN \\
        \{nrb, suc, nrb\} & temporal & eGCN \\
        \{rb, pre, rb\} & temporal & eGCN \\
        \{nrb, suc, nrb\} & temporal & eGCN \\
        \{rb, SSG, rb\} & spatial & GatV2 \\
        \{rb, SSG, nrb\} & spatial & GatV2 \\
        \{nrb, SSG, rb\} & spatial & GatV2 \\
        \{nrb, SSG, nrb\} & spatial & GatV2 \\
        \{m, suc-$i$, m\}, $i \in [1, 6]$ & spatial & eGCN \\
        \{m, pre-$i$, m\}, $i \in [1, 6]$ & spatial & eGCN \\
        \{m, right, m\} & spatial & eGCN \\
        \{m, left, m\} & spatial & eGCN \\
        \{rb, drives-on, m\} & spatial & GatV2 \\
        \{nrb, drives-on, m\} & spatial & GatV2 \\
        \{m, gives-traffic-info, rb\} & spatial & GatV2 \\
        \{m, gives-traffic-info, nrb\} & spatial & GatV2 \\
        \{rb, merge, rb\} & temporal & GatV2 \\
        \{nrb, merge, nrb\} & temporal & GatV2 \\
        \{map, anchor-$k$, rb\}, $k\in [0,10]$ & spatial & GatV2 \\
        \bottomrule
    \end{tabular}
    \label{tab:edges}
\end{table}
\subsection{Semantic Scene Graph}
\label{sec:ssg}
\begin{figure}[htbp]
  \centering
  \def\svgwidth{\columnwidth}
  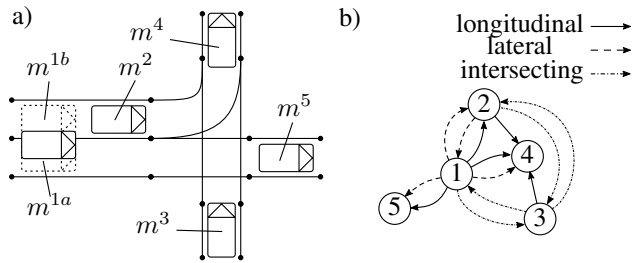
  \caption{Example of how five vehicles projected to near lanes result in six projection identities $m$ (a). The resulting semantic scene graph is presented in (b), where each traffic participant is represented by a node, and the edges represent the relations between its projection identities.\cite{zipfl_towards_2022}.}
  \label{fig:scene_graph}
\end{figure}

To utilize explicit social interaction types in the traffic, we employ a graph representation based on the semantic scene graph (SSG) developed in \cite{zipfl_towards_2022}.
The SSG focuses on topological information and discards a significant portion of geometric data.
In \Cref{fig:scene_graph}, we present an exemplary traffic scene (a) and its corresponding graph representation (b).
In the given graph, each traffic participant is represented as a node $v^i$, while the relationship between two participants ($i, j$) is defined as $e^{ij}$ based on their position on the road, considering the road topology and projection identity $m$.
In this manner, if two participants drive one after the other on the same lane, they will have a longitudinal relation %
(as shown for vehicle 2 and vehicle 4 in \Cref{fig:scene_graph}b).
In addition to %
the type of relationship, further information about the distance between them along the lane is stored.
Moreover, traffic participants may have lateral relations to adjacent entities or an intersection relation type if they travel on intersecting lanes.
We argue that the relation type and the distance of traffic participants along the lane and to a potential intersection/collision is an important factor that governs agent interaction and therefore should be considered in trajectory prediction.
One example could be two vehicles in an intersection. Having a longitudinal relations means one vehicle is just following the other. An intersecting relation indicates that the paths of the vehicles are intersecting with a potential collision resulting in a different social relation.
\subsection{Map Autoencoder}
\label{sec:map_encoding}
In addition to the lane centerlines represented by the map nodes, HD-Maps like the nuScenes map also include other relevant information for prediction like lane borders, lane dividers, pedestrian crossings, walkways, traffic signs and stop areas.
We consider this important information that influences the behavior of traffic participants and therefore should be included in the prediction approach.
In the nuScenes dataset \cite{caesar_nuscenes_2020}, which is used in this work, this information is provided in the form of 10 binary masks, one for each of the different types.
To augment the road geometry with this additional information, we use an map autoencoder (MA) to compress the 10 channel images in a latent representation $z_\text{map}$.
We opt for a simple autoencoder architecture that reduces the $10x128x128$ input to a hidden dimension of $128$.
The encoder consists of six layers, each consisting of a 2D-Convolution, Batch Normalization and LeakyReLU activation.
The decoder has also six layers, where the first five layers consist of 2D-Deconvolution, Batch Normalization and ReLU activation.
In the last layer, we use tanh as activation function.
The $128x128$ input image covers a spatial distance of $50~m$ x $50~m$.
We trained the autoencoder for 100 epochs with a learning rate of $2\cdot 10^{-4}$ and reached a MSE reconstruction loss of $0.59$. 
We extend rb-agent-nodes and nrb-agent-nodes with additional information from the map.
Therefore, we add the latent representation of the local patch $z_\text{map}$ of the environment around a node to the features of the node before the Merge Stage, see \Cref{fig:arcitecture}:
\begin{equation}
    {}^{\{rb,nrb\}}n_{i}^{t} = \text{MLP}\left(\text{concat}\left({}^{\{rb,nrb\}}n_{i}^{t}, \text{dropout}\left(z_{\text{map}_i}^{t}\right)\right)\right)
\end{equation}
To overcome Overfitting, we use a dropout rate of $0.5$.
The patch is centered around the agent of interest and rotated so that the direction of movement is facing north.
\subsection{Anchor Paths}
\label{sec:ap}
This section describes the usage of anchor paths (AP) to force prediction results to stay within permitted trajectory paths.
APs and the Lanelet2 \cite{poggenhans_lanelet2_2018} conversion of the nuScenes map are generated by a framework proposed in  \cite{naumannHertlein2023lanelet2ForNuScenes}.
Lanelet2 \cite{poggenhans_lanelet2_2018} is an open-source framework to generate and work with HD-Maps.
To generate APs the rb-agents are first mapped to lanelets, then lane-anchors are generated, that consist of a list of lane ids representing permitted paths.
We define lane-anchors to have a maximal length of $100~m$ and describe all possible paths an rb-agent can take including potential lane changes.
Lane-anchors are used to generate APs and anchor-embeddings (AEs).
An AP is defined as a polyline with the 2d-positions.
An AE $z_\text{anchor}$ is a latent representation of the anchor with a size of $128$.
AEs are generated by the GNN via the edges \{map, anchor-k, rb\}, with $k \in [0,9]$.
Those edges do not update the rb-agent-nodes directly.
Instead, each specific anchor edge-type generates an anchor embedding $z_{\text{anchor},i,k}$.
In the trajectory decoder, the rb-agent-features are copied k times.
Afterwards, they are concatenated with the anchor-embeddings and used by the prediction head to output scored trajectories for road-bound agents ${}^{rb}\pmb{\mathcal{T}}$ and non-road-bound agents ${}^{nrb}\pmb{\mathcal{T}}$.
\begin{equation}
   {}^{rb}\pmb{\mathcal{T}}_k = {}^{rb}\text{MLP}_{k}(\text{concat}({}^{rb}z_{\text{agent}_{i,k}}, z_{\text{anchor}_{i,k}}))
\end{equation}
\begin{equation}
   {}^{nrb}\pmb{\mathcal{T}}_k = {}^{nrb}\text{MLP}_{k}({}^{nrb}z_{\text{agent}_{i,k}})
\end{equation}
\subsection{Loss}
\label{seq:loss}
The Loss $\mathcal{L}$ is defined in \Cref{eq:loss}.
\begin{equation}
    \mathcal{L} = \mathcal{L}_\text{reg} + w_1 \mathcal{L}_\text{score} + w_2 \mathcal{L}_\text{yaw}
    \label{eq:loss}
\end{equation}
The regression loss $\mathcal{L}_\text{ref}$ which is a smooth L1 loss and the scoring loss $\mathcal{L}_\text{score}$ which is a max-margin loss are only evaluated on the predicted trajectory with the least final displacement error to the ground truth trajectory.
The orientation loss $\mathcal{L}_\text{yaw}$ 
is evaluated only for road-bound agents.
By comparing the orientation of the predicted trajectories with the orientation of the APs, the model should be forced to output trajectories that tend to follow the APs.
In order to consider that angles are defined in $(-\pi, \pi]$ and that similar angles can be expressed as a multiple of $2\pi$, the orientation loss for an agent $i$ in the mode $k$ is defined as:
\begin{equation}
    \mathcal{L}_{\text{yaw}_{i,k}} = 1 - \text{cos}(\theta_{\text{anchor}_{i,k}} - \theta_{i,k})
\end{equation}
We set the parameters to $w_1 = 1$ and $w_2 = 1$.
Loss terms are compressed over all agents in the batch using mean.

\section{Evaluation}
\label{sec:evaluation}
In this section, we evaluate our model and demonstrate the meaningfulness of our contribution through an ablation study on the nuScenes dataset.

\subsection{Experimental Settings}
The nuScenes dataset is a large collection of traffic scenarios that are collected in Boston and Singapore.
For the prediction, we use the official train/val split with 32186/9041 samples.
Each sample contains trajectories, sampled at $2~Hz$, of all agents in the scene.
$2~s$ of history is given and the task is to predict the trajectory for the next $6~s$ at $2~Hz$ of a specific agent in the scene.
The HD-Map data is provided in a nuScenes specific format, however we use the HD-Map in a lanelet specific format proposed by \cite{naumannHertlein2023lanelet2ForNuScenes}.
We use the metrics proposed in the nuScenes prediction challenge\footnote{https://www.nuscenes.org/prediction}:
\begin{itemize}
    \item $\text{minADE}_\text{k}$ - The average of pointwise L2 distances between the predicted trajectory and ground truth over the k most likely predictions.
    \item $\text{minFDE}_\text{k}$ - L2 distance between the final point of the predicted trajectory and ground truth over the k most likely predictions.
    \item $\text{MR}_\text{2,k}$ - If the maximum pointwise L2 distance between the prediction and ground truth is greater than 2 meters, it is defined as a miss. For each agent, the k most likely predictions are taken and evaluated if any are misses. The $\text{MR}_\text{2,k}$ is the proportion of misses over all agents.
    \item $\text{ORR}$ - The OffRoadRate(ORR) is defined as the fraction of predicted trajectories that are not entirely contained in the drivable area of the map.
\end{itemize}
We use a Nvidia RTX 3080 GPU to train the model for 40 epochs with Adam Optimizer \cite{kingma_adam_2017}, an initial learning rate of $10^{-3}$ and a learning rate decay of $0.5$ every fifth epoch.
Further, we use an accumulated batch size of 64 and a weight decay of $0.5~\%$.
The attention layers use $4$ heads and concatenate the results of each head.
We set the origin of the local coordinate system to be in the geometric center of each sample.
Map-nodes are gathered in a square of size $190~m$ x $190~m$.
\subsection{Ablation Study}
\begin{table*}[t]
    \centering
    \caption{Results on nuScenes validation split in comparison to state-of-the-art models}
    \begin{tabular}{@{}l  c c c c c c@{}}
         \toprule
         Model & $\text{minADE}_{5}$ & $\text{minADE}_{10}$ & $\text{minFDE}_{1}$ & $\text{minMR}_{5}$ & $\text{minMR}_{10}$ & ORR \\ \midrule
         CoverNet \cite{phan-minh_covernet_2020} & 1.96~m & 1.48~m & 9.26~m & 0.67 & - & - \\
         Trajectron++ \cite{salzmann_trajectron_2021} & 1.88~m & 1.48~m & 9.52~m & 0.70 & 0.57 & 0.25 \\
         HG \cite{holigraph} & 1.82~m & 1.38~m & 9.61~m & 0.68 & 0.63 & 0.12 \\
         CXX \cite{luo_probabilistic_2020} & 1.63~m & 1.29~m & 8.86~m & 0.69 & 0.60 & 0.08 \\
         MultiPath \cite{varadarajan_multipath_2021} & 1.78~m & 1.55~m & 10.16~m & - & - & - \\
         PGP \cite{deo_multimodal_2021} & 1.30~m & 1.00~m & 7.17~m & 0.61 & 0.37 & 0.03 \\
         Thomas \cite{Gilles2021THOMASTH} & 1.33~m & 1.04~m & 6.71~m & 0.55 & 0.42 & 0.03 \\
         HG+SSG+AP+MA &  1.63~m & 1.29~m & 9.29~m & 0.65 & 0.57 & 0.07\\
         \bottomrule
    \end{tabular}
    \label{tab:results}
\end{table*}
We conducted an extensive ablation study to show the effect of all our contributions.
As a baseline, we used the initial HoliGraph (HG) approach \cite{holigraph}.
Firstly, we show the effect of differentiating between the two agent types \emph{road-bound} and \emph{non-road-bound}.
For the nuScenes dataset, the classes car, bus, truck, trailer, ambulance, police and motorcyclist were included to type \emph{road-bound}. 
Adult, construction worker, child, wheelchair, police officer and cyclist belonged to \emph{non-road-bound}.
Therefore, we evaluated the model on all agents in the nuScenes validation split.
Be aware, that the results in \Cref{tab:ablation1} are not comparable to the nuScenes leaderboard, because the nuScenes challenge defines one agent in a scene as target and ignores the prediction of all others.
The distinction between \emph{road-bound} and \emph{non-road-bound} improves the performance of the model for \emph{non-road-bound} agents and can be explained by looking at the distribution of agent-types across the dataset.
In total, there are 686,038 agents in the nuScenes train split.
485,487 (71\%) of which are \emph{road-bound} agents and 200,551 (29\%) are \emph{non-road-bound} agents.
\emph{Non-road-bound} agents are underrepresented in the training dataset, hence the baseline focuses mainly on \emph{road-bound} agents.
Handling the class types with different types of nodes helps to diminish this shortcoming and is used in the following experiments.
\begin{table}[h]
    \centering
    \caption{Distinguishing between \emph{road-bound} and \emph{non-road-bound} agents. Metrics are calculated \textbf{on all agents} in the nuScenes validation split. Agent splitting decreases performance for \emph{road-bound} agents slightly but improves performance for \emph{non-road-bound} agents considerably.}
    \begin{tabular}{@{}l  c c c c@{}} \toprule
         & split agent-types & $\text{minADE}_{10}$ & $\text{minFDE}_{10}$ & $\text{minMR}_{10}$ \\ \midrule
         rb &  & 0.64~m & 1.08~m & 0.19 \\ 
         nrb &  & 0.47~m & 0.90~m & 0.10 \\ \midrule
         rb & \multirowcell{2}{\checkmark} & 0.66~m & 1.11~m & 0.20 \\ 
         nrb & & 0.38~m & 0.66~m & 0.05 \\ \bottomrule 
    \end{tabular}
    \label{tab:ablation1}
\end{table}
\begin{figure*}[t]
    \centering
    \vspace{10pt}
    \begin{subfigure}{0.33\textwidth}
        \centering
        \includegraphics[width=\columnwidth]{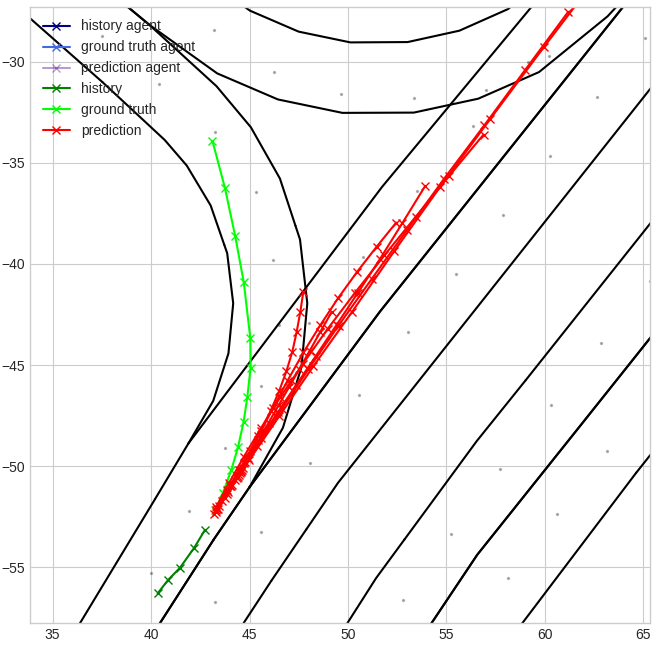}
        \caption{Mode collapse in HG \cite{holigraph}.}
        \label{fig:mode_collapse}
    \end{subfigure}
    \hspace{30pt}
    \begin{subfigure}{0.33\textwidth}
        \centering
        \includegraphics[width=\columnwidth]{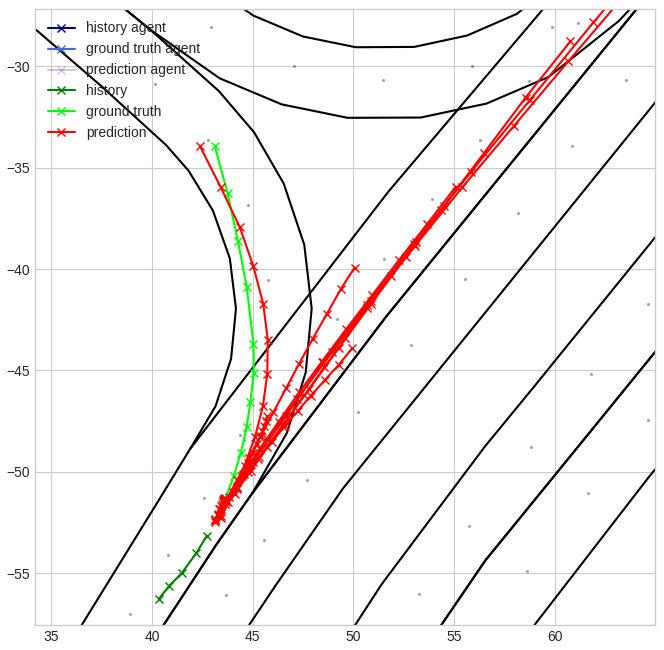}
        \caption{No Mode collapse in our model.}
        \label{fig:no_mode_collapse}
    \end{subfigure}
    \caption{Comparison between the HG baseline and our model to show, that our model successfully overcomes mode collapse.}
    \label{fig:mode_collapse_main}
\end{figure*}
\begin{table}[ht]
    \hspace{0.1pt}
    \centering
    \caption{Ablation study on the nuScenes validation split that demonstrates the influence of our contributions. Metrics \textbf{are only calculated on target agents} marked by the dataset. Relative improvement is calculated between first and last row.}
    \begin{tabular}{>{\centering}m{0.6cm}
                    >{\centering}m{0.4cm}
                    >{\centering}m{0.6cm} 
                    >{\centering}m{1.0cm}
                    >{\centering}m{1.0cm}
                    >{\centering}m{1.0cm}
                    >{\centering\arraybackslash}m{0.9cm}}
        \toprule
        SSG & AP & MA & $\text{minADE}_{10}$ & $\text{minFDE}_{10}$ & $\text{minMR}_{10}$ & ORR \\ \midrule
         & & & \textbf{1.38~m} & \textbf{2.60~m} & \textbf{0.63} & \textbf{0.12}\\
        \checkmark & & & 1.40~m & 2.61~m & 0.63 & 0.12 \\
        \checkmark & \checkmark & & 1.31~m & 2.43~m & 0.60 & 0.08 \\
        \checkmark &  & \checkmark & 1.35~m & 2.52~m & 0.62 & 0.10 \\
        \checkmark & \checkmark & \checkmark & \textbf{1.29~m} & \textbf{2.32~m} & \textbf{0.57} & \textbf{0.06} \\ \midrule
        \multicolumn{3}{c}{relative improvement} & 6\% & 10\% & 9\% & 50\%\\ \bottomrule
    \end{tabular}
    \label{tab:ablation2}
\end{table}

Next, we performed experiments where we evaluated the three additional features: Semantic Scene Graph (SSG), Anchor Paths (AP), Map Autoencoder (MA).
This time the evaluation is only done on agents, marked as target by the nuScenes prediction challenge, i.e., a single agent in each scene.
We start by adding the SSG features to the baseline.
The \emph{social}-edges of the baseline, which are defined in a fully connected fashion between every agent-related node from the same time-step, were exchanged with the \emph{SSG}-edges that describe the relation type and distance between the agents along the lane.
Results indicate that these features do not improve prediction performance.
We consider the relation type "interacting", which describes two agents that could interact, as important information for prediction because it often indicates that one of the agents has to yield to the other agent at an intersection.
Since the nuScenes dataset mainly contains intersections that are governed by traffic lights where the state of the traffic light is not provided, this information cannot be directly considered in the subsequent modeling process. 
We hypothesise this the reason for no performance improvements using SSG information.
Comparing the average number of \emph{social}-edges in a graph with the average number of \emph{SSG}-edges in a graph shows that using the SSG to describe social interaction reduces the number of respective necessary edges by $90\%$ on nuScenes.
This results in reduced complexity and size of the graph without decreasing performance. Therefore, we argue that using the scenario graph for social interaction is beneficial for motion prediction in AVs.

Then we performed experiments using AP and MA features separately and together on top of the SSG features.
Using AP for \emph{road-bound} agents as well as using an MA improve trajectory prediction performance.
Using AP and MA features together with SSG features leads to the best overall trajectory prediction performance increasing performance on all metrics, see \Cref{tab:ablation2}.

A common problem in multimodal motion prediction is mode collapse, i.e., all predicted trajectories follow the same driving behaviour.
This shortcoming is particularly severe when predicting agents near an intersection.
\Cref{fig:mode_collapse} shows the predictions of the baseline model.
The potential left turn is completely neglected by the model.
The usage of AP helps to avoid mode collapse, see \Cref{fig:no_mode_collapse}.
\subsection{Comparison with other methods}
Although our goal is to introduce and validate additional features that alleviate shortcomings of current motion prediction methods which we evaluated in the previous chapter on the HG approach, for completeness we also provide a comparison of our approach with other state-of-the-art methods.
However, it should be noted that our baseline using HG is a relatively simple approach compared to the best-performing methods shown below.
Based on the obtained performance improvements with our proposed three features we also expect to obtain improved performance when these three features are combined with other top-performing methods.
Results on nuScenes validation split are shown in \Cref{tab:results}, indicating that our model matches the performance of the state-of-the-art.
However, other models, like PGP, perform better on the validation set while having similar ORR to our approach.
It can be observed that all agents used for the metrics are \emph{road-bound}, e.g., vehicles or trucks, and that all models have similar low ORR.
Thus, the metric results mainly depend on the correct prediction of the longitudinal displacement.
PGP reports, they are using a special velocity sampling where they predict 200 trajectories and cluster them into 10 predictions to better handle the task.
We assume that this strategy has a high impact on the longitudinal displacement and that our model would also benefit from that.
The model provided in our work has not the goal to outperform the state-of-the-art but instead showcases beneficial methods to improve existing models, such as HG \cite{holigraph}, but also other models like HiVT \cite{zhou_hivt_2022} and PGP \cite{deo_multimodal_2021} could be improved.

\section{Conclusion}
\label{sec:conclustion}
We propose three building blocks to address the shortcomings of state-of-the-art motion prediction methods. These consist of a) a semantic scene graph (SSG) that is able to differentiate between different relation types between agents as well as their distance along the lane, b) an agent-centric autoencoder-based map representation (MA) able to encode local map context, and c) anchor paths (AP) that guide trajectory prediction results to stay within permitted driving paths.
Our evaluation shows that AP and MA improved the performance considerably.
This proves, that MA makes more information about the static environment available to the model and AP acts as a seed in the decoder to predict more diverse trajectories in compliance to the road topology.
The SSG did not improve performance which could be due to the predominance of traffic light controlled intersections in the dataset, which cannot be accounted for in the SSG.
Nevertheless, the SSG reduced the complexity of the heterogeneous graph by decreasing the number of edges for social interaction between traffic participants by up to $90\%$.
The improved trajectory prediction results based on the proposed approach suggest that also other state-of-the-art methods could be improved with this approach.

\bibliographystyle{IEEEtran}

\bibliography{references_holigraph2, references_offline, references_Juergen_2}

\end{document}